\documentclass[11pt,a4paper]{article}
\usepackage[hyperref]{naaclhlt2018}
\usepackage{times}
\usepackage{latexsym}

\usepackage{amsmath,graphicx}
\usepackage{amsfonts}
\usepackage{algorithm}
\usepackage{algpseudocode}

\usepackage{url}
  
\aclfinalcopy 

\algnewcommand{\parState}[1]{\State%
  \parbox[t]{\dimexpr\linewidth-\algmargin}{\strut #1\strut}}

\title{Dialogue Learning with Human Teaching and Feedback in End-to-End Trainable Task-Oriented Dialogue Systems}

\author{Bing Liu$^1$\thanks{\hspace{2mm}Work done while the author was an intern at Google.}, Gokhan T\"{u}r$^2$, Dilek Hakkani-T\"{u}r$^2$, Pararth Shah$^2$, Larry Heck$^3$\thanks{\hspace{2mm}Work done while at Google Research.} \\
  $^1$Carnegie Mellon University, Pittsburgh, PA, USA \\
  $^2$Google Research, Mountain View, CA,USA
  $^3$Samsung Research, Mountain View, CA, USA \\
  {\tt liubing@cmu.edu}, {\tt \{dilekh,pararth\}@google.com}, \\
  {\tt \{gokhan.tur,larry.heck\}@ieee.org}\\
   \\}

\date{}

\begin{document}
\maketitle
\begin{abstract}
    In this work, we present a hybrid learning method for training task-oriented dialogue systems through online user interactions. Popular methods for learning task-oriented dialogues include applying reinforcement learning with user feedback on supervised pre-training models. Efficiency of such learning method may suffer from the mismatch of dialogue state distribution between offline training and online interactive learning stages. To address this challenge, we propose a hybrid imitation and reinforcement learning method, with which a dialogue agent can effectively learn from its interaction with users by learning from human teaching and feedback. We design a neural network based task-oriented dialogue agent that can be optimized end-to-end with the proposed learning method. Experimental results show that our end-to-end dialogue agent can learn effectively from the mistake it makes via imitation learning from user teaching. Applying reinforcement learning with user feedback after the imitation learning stage further improves the agent's capability in successfully completing a task.
\end{abstract}

\section{Introduction}

\label{sec:intro}
    Task-oriented dialogue systems assist users to complete tasks in specific domains by understanding user's request and aggregate useful information from external resources within several dialogue turns. Conventional task-oriented dialogue systems have a complex pipeline~\cite{rudnicky1999creating,raux2005let,young2013pomdp} consisting of independently developed and modularly connected components for natural language understanding (NLU)~\cite{mesnil2015using,liu2016Joint,hakkani2016multi}, dialogue state tracking (DST)~\cite{henderson2014word,mrkvsic2016neural}, and dialogue policy learning~\cite{gasic2014gaussian,shah2016interactive,su2016line,su2017sample}. These system components are usually trained independently, and their optimization targets may not fully align with the overall system evaluation criteria (e.g. task success rate and user satisfaction). Moreover, errors made in the upper stream modules of the pipeline propagate to downstream components and get amplified, making it hard to track the source of errors. 
    
    To address these limitations with the conventional task-oriented dialogue systems, recent efforts have been made in designing end-to-end learning solutions with neural network based methods. Both supervised learning (SL) based~\cite{wenN2N16,bordes2017,Liu2017} and deep reinforcement learning (RL) based systems~\cite{zhao2016towards,li2017end,peng2017composite} have been studied in the literature. Comparing to chit-chat dialogue models that are usually trained offline using single-turn context-response pairs, task-oriented dialogue model involves reasoning and planning over multiple dialogue turns. This makes it especially important for a system to be able to learn from users in an interactive manner. Comparing to SL models, systems trained with RL by receiving feedback during users interactions showed improved model robustness against diverse dialogue scenarios~\cite{williams2016end,liu2017iterative}. 
    
    A critical step in learning RL based task-oriented dialogue models is dialogue policy learning. Training dialogue policy online from scratch typically requires a large number of interactive learning sessions before an agent can reach a satisfactory performance level. Recent works~\cite{henderson2008hybrid,williams2017hybrid,liu2017e2e} explored pre-training the dialogue model using human-human or human-machine dialogue corpora before performing interactive learning with RL to address this concern. A potential drawback with such pre-training approach is that the model may suffer from the mismatch of dialogue state distributions between supervised training and interactive learning stages. While interacting with users, the agent's response at each turn has a direct influence on the distribution of dialogue state that the agent will operate on in the upcoming dialogue turns. If the agent makes a small mistake and reaches an unfamiliar state, it may not know how to recover from it and get back to a normal dialogue trajectory. This is because such recovery situation may be rare for good human agents and thus are not well covered in the supervised training corpus. This will result in compounding errors in a dialogue which may lead to failure of a task. RL exploration might finally help to find corresponding actions to recover from a bad state, but the search process can be very inefficient. 
    
    To ameliorate the effect of dialogue state distribution mismatch between offline training and RL interactive learning, we propose a hybrid imitation and reinforcement learning method. We first let the agent to interact with users using its own policy learned from supervised pre-training. When an agent makes a mistake, we ask users to correct the mistake by demonstrating the agent the right actions to take at each turn. This user corrected dialogue sample, which is guided by the agent's own policy, is then added to the existing training corpus. We fine-tune the dialogue policy with this dialogue sample aggregation~\cite{ross2011reduction} and continue such user teaching process for a number of cycles. Since asking for user teaching at each dialogue turn is costly, we want to reduce this user teaching cycles as much as possible and continue the learning process with RL by collecting simple forms of user feedback (e.g. a binary feedback, positive or negative) only at the end of a dialogue. 

    Our main contributions in this work are:
    \begin{itemize}
        \item We design a neural network based task-oriented dialogue system which can be optimized end-to-end for natural language understanding, dialogue state tracking, and dialogue policy learning.
        \item We propose a hybrid imitation and reinforcement learning method for end-to-end model training in addressing the challenge with dialogue state distribution mismatch between offline training and interactive learning.
    \end{itemize}
    
    The remainder of the paper is organized as follows. In section 2, we discuss related work in building end-to-end task-oriented dialogue systems. In section 3, we describe the proposed model and learning method in detail. In Section 4, we describe the experiment setup and discuss the results. Section 5 gives the conclusions. 

\section{Related Work}
\label{sec:related_work}
    Popular approaches in learning task-oriented dialogue include modeling the task as a partially observable Markov Decision Process (POMDP)~\cite{young2013pomdp}. RL can be applied in the POMDP framework to learn dialogue policy online by interacting with users~\cite{gavsic2013line}. The dialogue state and system action space have to be carefully designed in order to make the policy learning tractable~\cite{young2013pomdp}, which limits the model's usage to restricted domains. 

    Recent efforts have been made in designing end-to-end solutions for task-oriented dialogues, inspired by the success of encoder-decoder based neural network models in non-task-oriented conversational systems~\cite{serban2015building,li2016persona}. Wen et al. \cite{wenN2N16} designed an end-to-end trainable neural dialogue model with modularly connected system components. This system is a supervised learning model which is evaluated on fixed dialogue corpora. It is unknown how well the model performance generalizes to unseen dialogue state during user interactions. Our system is trained by a combination of supervised and deep RL methods, as it is shown that RL may effectively improve dialogue success rate by exploring a large dialogue action space~\cite{henderson2008hybrid,li2017end}.  

    Bordes and Weston~\shortcite{bordes2017} proposed a task-oriented dialogue model using end-to-end memory networks. In the same line of research, people explored using query-regression networks~\cite{seo2016query}, gated memory networks~\cite{liu2017gated}, and copy-augmented networks~\cite{eric2017copy} to learn the dialogue state. These systems directly select a final response from a list of response candidates conditioning on the dialogue history without doing slot filling or user goal tracking. Our model, on the other hand, explicitly tracks user's goal for effective integration with knowledge bases (KBs). Robust dialogue state tracking has been shown~\cite{jurvcivcek2012reinforcement} to be critical in improving dialogue success in task completion.

  \begin{figure*}[t]
      \centering
      \includegraphics[width=420pt]{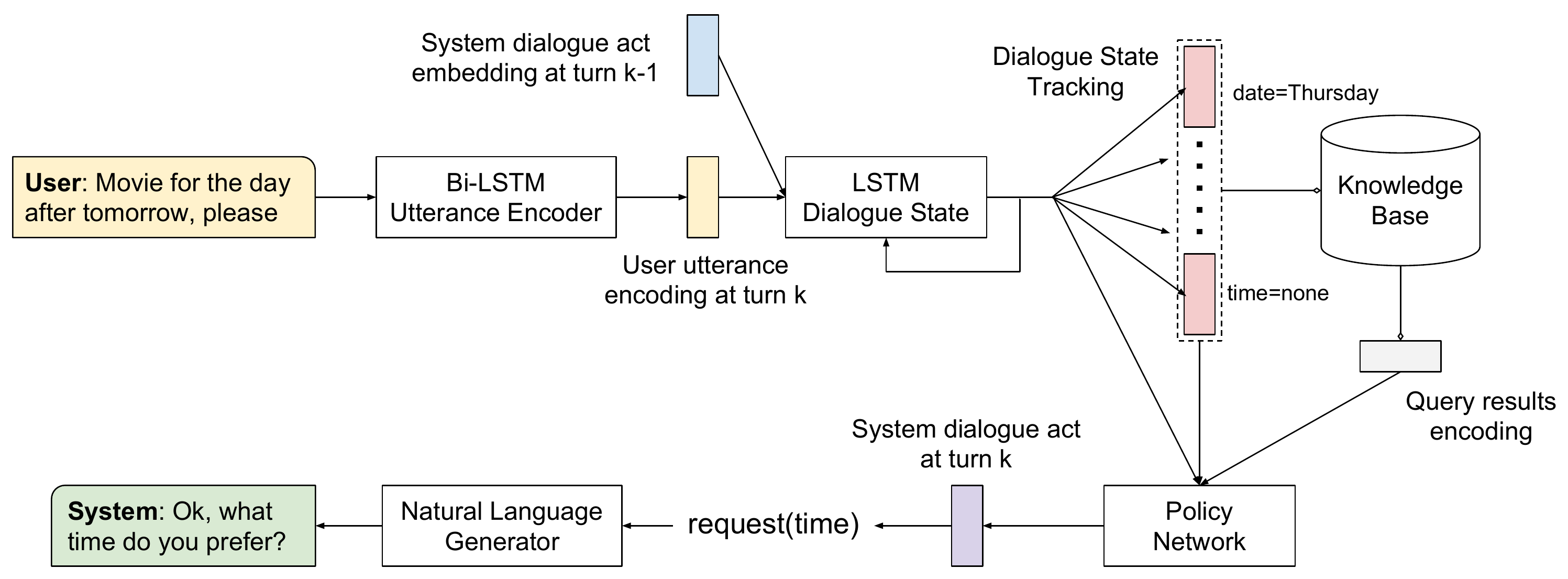}
      \caption{{ Proposed end-to-end task-oriented dialogue system architecture. }}
      \label{fig:system_design}
  \end{figure*}
  
    Dhingra et al.~\shortcite{dhingra2017towards} proposed an end-to-end RL dialogue agent for information access. Their model focuses on bringing differentiability to the KB query operation by introducing a ``soft'' retrieval process in selecting the KB entries. Such soft-KB lookup is prone to entity updates and additions in the KB, which is common in real world information systems. In our model, we use symbolic queries and leave the selection of KB entities to external services (e.g. a recommender system), as entity ranking in real world systems can be made with much richer features (e.g. user profiles, location and time context, etc.). Quality of the generated symbolic query is directly related to the belief tracking performance. In our proposed end-to-end system, belief tracking can be optimized together with other system components (e.g. language understanding and policy) during interactive learning with users.

    Williams et al.~\shortcite{williams2017hybrid} proposed a hybrid code network for task-oriented dialogue that can be trained with supervised and reinforcement learning. They show that RL performed with a supervised pre-training model using labeled dialogues improves learning speed dramatically. They did not discuss the potential issue of dialogue state distribution mismatch between supervised pre-training and RL interactive learning, which is addressed in our dialogue learning framework.

\label{sec:method}
\section{Proposed Method}
    Figure \ref{fig:system_design} shows the overall system architecture of the proposed end-to-end task-oriented dialogue model. We use a hierarchical LSTM neural network to encode a dialogue with a sequence of turns. User input to the system in natural language format is encoded to a continuous vector via a bidirectional LSTM utterance encoder. This user utterance encoding, together with the encoding of the previous system action, serves as the input to a dialogue-level LSTM. State of this dialogue-level LSTM maintains a continuous representation of the dialogue state. Based on this state, the model generates a probability distribution over candidate values for each of the tracked goal slots. A query command can then be formulated with the state tracking outputs and issued to a knowledge base to retrieve requested information. Finally, the system produces a dialogue action, which is conditioned on information from the dialogue state, the estimated user's goal, and the encoding of the query results . This dialogue action, together with the user goal tracking results and the query results, is used to generate the final natural language system response via a natural language generator (NLG). We describe each core model component in detail in the following sections.

\subsection{Utterance Encoding}
    We use a bidirectional LSTM to encode the user utterance to a continuous representation. We refer to this LSTM as the utterance-level LSTM. The user utterance vector is generated by concatenating the last forward and backward LSTM states. Let $\mathbf{U}_k = (w_1, w_2, ..., w_{T_k})$ be the user utterance at turn $k$ with $T_k$ words. These words are firstly mapped to an embedding space, and further serve as the step inputs to the bidirectional LSTM. Let $\overrightarrow{h_{t}}$ and $\overleftarrow{h_{t}}$ represent the forward and backward LSTM state outputs at time step $t$. The user utterance vector $U_k$ is produced by: $U_k =  [\overrightarrow{h_{T_k}}, \overleftarrow{h_{1}}]$, where $\overrightarrow{h_{T_k}}$ and $\overleftarrow{h_{1}}$ are the last states in the forward and backward LSTMs.
    
\subsection{Dialogue State Tracking}
    Dialogue state tracking, or belief tracking, maintains the state of a conversation, such as user's goals, by accumulating evidence along the sequence of dialogue turns. Our model maintains the dialogue state in a continuous form in the dialogue-level LSTM ($\operatorname{LSTM_D}$) state $s_k$. $s_k$ is updated after the model processes each dialogue turn by taking in the encoding of user utterance $U_k$ and the encoding of the previous turn system output $A_{k-1}$. This dialogue state serves as the input to the dialogue state tracker. The tracker updates its estimation of the user's goal represented by a list of slot-value pairs. A probability distribution $P(l^{m}_k)$ is maintained over candidate values for each goal slot type $m \in M$: 
        \begin{align}
            & s_k = \operatorname{LSTM_D}(s_{k-1}, \hspace{1mm} [U_k, \hspace{1mm} A_{k-1}]) \\
            & P(l^{m}_k \hspace{1mm} | \hspace{1mm} \mathbf{U}_{\le k}, \hspace{1mm} \mathbf{A}_{< k}) = \operatorname{SlotDist}_{m}(s_k)
        \end{align}
    where $\operatorname{SlotDist}_{m}$ is a single hidden layer MLP with $\operatorname{softmax}$ activation over slot type $m \in M$.

\subsection{KB Operation}
    The dialogue state tracking outputs are used to form an API call command to retrieve information from a knowledge base. The API call command is produced by replacing the tokens in a query command template with the best hypothesis for each goal slot from the dialogue state tracking output. Alternatively, an n-best list of API calls can be generated with the most probable candidate values for the tracked goal slots. 
    In interfacing with KBs, instead of using a soft KB lookup as in~\cite{dhingra2017towards}, our model sends symbolic queries to the KB and leaves the ranking of the KB entities to an external recommender system. Entity ranking in real world systems can be made with much richer features (e.g. user profiles, local context, etc.) in the back-end system other than just following entity posterior probabilities conditioning on a user utterance. Hence ranking of the KB entities is not a part of our proposed neural dialogue model. In this work, we assume that the model receives a ranked list of KB entities according to the issued query and other available sources, such as user models.
    
    Once the KB query results are returned, we save the retrieved entities to a queue and encode the result summary to a vector. Rather then encoding the real KB entity values as in~\cite{bordes2017,eric2017copy}, we only encode a summary of the query results (i.e. item availability and number of matched items). This encoding serves as a part of the input to the policy network.
    
\subsection{Dialogue Policy}
    A dialogue policy selects the next system action in response to the user's input based on the current dialogue state. We use a deep neural network to model the dialogue policy. There are three inputs to the policy network, (1) the dialogue-level LSTM state $s_{k}$, (2) the log probabilities of candidate values from the belief tracker $v_{k}$, and (3) the encoding of the query results summary $E_{k}$. The policy network emits a system action in the form of a dialogue act conditioning on these inputs:
    \begin{align}
        P(a_{k} \hspace{1mm} | \hspace{1mm} U_{\le k}, \hspace{1mm} A_{< k}, \hspace{1mm} E_{\le k}) = \operatorname{PolicyNet}(s_{k}, v_{k}, E_{k})
    \end{align}
    where $v_{k}$ represents the concatenated log probabilities of candidate values for each goal slot, $E_{k}$ is the encoding of query results, and $\operatorname{PolicyNet}$ is a single hidden layer MLP with $\operatorname{softmax}$ activation function over all system actions.
    \begin{figure}[t]
        \centering
        \includegraphics[width=200pt]{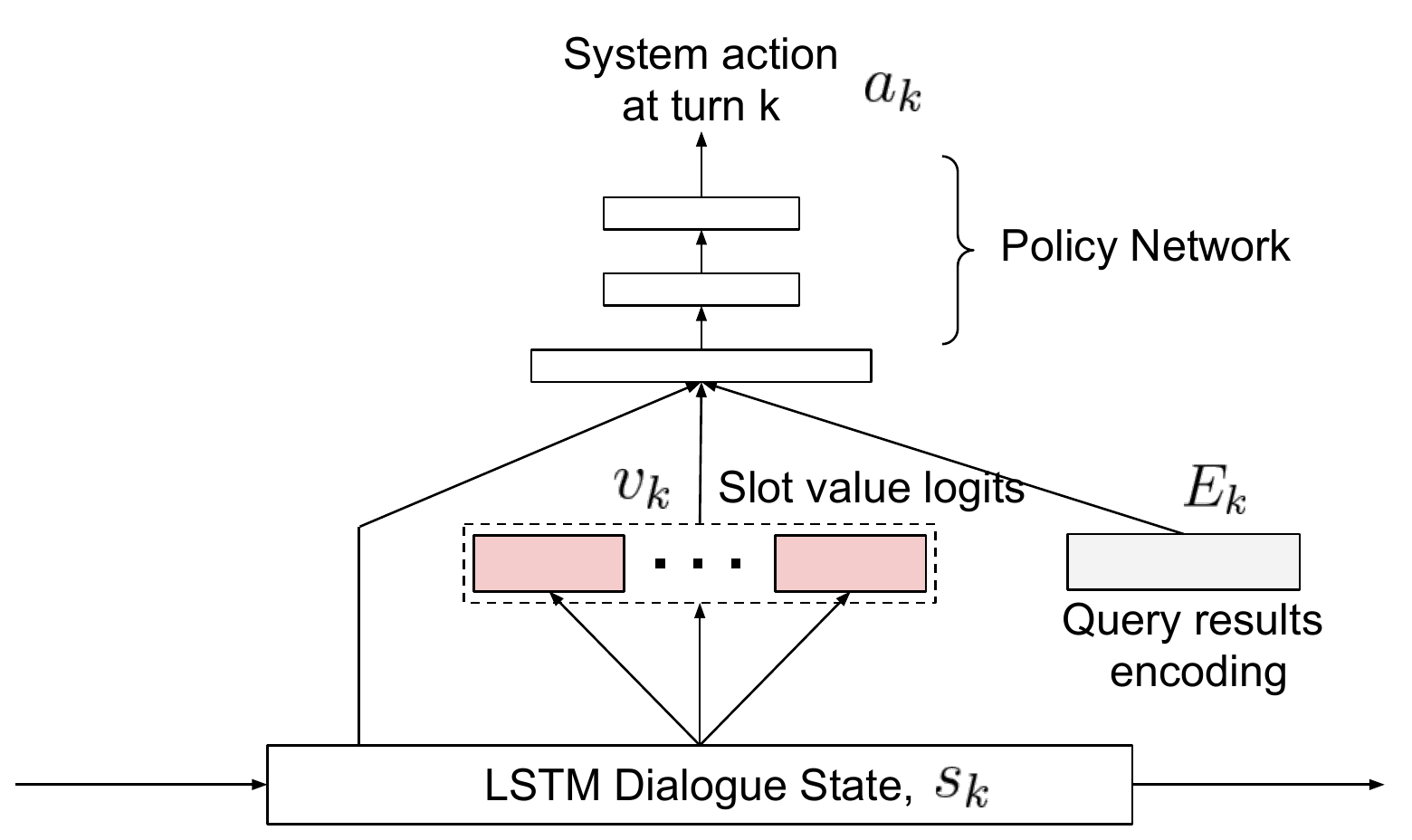}
        \caption{{ Dialogue state and policy network. }}
        \label{fig:policy_network}
    \end{figure}

    The emitted system action is finally used to produce a system response in natural language format by combining the state tracker outputs and the retrieved KB entities. We use a template based NLG in this work. The delexicalised tokens in the NLG template are replaced by the values from either the estimated user goal values or the KB entities, depending on the emitted system action. 

\subsection{Supervised Pre-training}
    By connecting all the system components, we have an end-to-end model for task-oriented dialogue. Each system component is a neural network that takes in underlying system component's outputs in a continuous form that is fully differentiable, and the entire system (utterance encoding, dialogue state tracking, and policy network) can be trained end-to-end. 
    
    We first train the system in a supervised manner by fitting task-oriented dialogue samples. The model predicts the true user goal slot values and the next system action at each turn of a dialogue. We optimize the model parameter set $\theta$ by minimizing a linear interpolation of cross-entropy losses for dialogue state tracking and system action prediction:
        \begin{equation}
            \begin{split}
            \min_{\theta} \sum_{k=1}^{K} -\Big[ \sum_{m=1}^{M} &\lambda _{l^{m}} \log P({l^{m}_k}^{*} | \mathbf{U}_{\le k}, \mathbf{A}_{< k}, \mathbf{E}_{< k}; \theta) \\
            + &\lambda _a \log P(a_k^{*} | \mathbf{U}_{\le k}, \mathbf{A}_{< k}, \mathbf{E}_{\le k}; \theta) \hspace{1mm} \Big]
            \end{split}
        \end{equation}
    where $\lambda$s are the linear interpolation weights for the cost of each system output. ${l^{m}_k}^{*}$ is the ground truth label for the tracked user goal slot type $m \in M$ at the $k$th turn, and $a_k^{*}$ is the true system action in the corpus.

\subsection{Imitation Learning with Human Teaching}
    Once obtaining a supervised training dialogue agent, we further let the agent to learn interactively from users by conducting task-oriented dialogues. Supervised learning succeeds when training and test data distributions match. During the agent's interaction with users, any mistake made by the agent or any deviation in the user's behavior may lead to a different dialogue state distribution than the one that the supervised learning agent saw during offline training. A small mistake made by the agent due to this covariate shift~\cite{ross2010efficient,ross2011reduction} may lead to compounding errors which finally lead to failure of a task. To address this issue, we propose a dialogue imitation learning method which allows the dialogue agent to learn from human teaching. We let the supervised training agent to interact with users using its learned dialogue policy $\pi _{\theta} (a|s)$. With this, we collect additional dialogue samples that are guided by the agent's own policy, rather than by the expert policy as those in the supervised training corpora. When the agent make mistakes, we ask users to correct the mistakes and demonstrate the expected actions and predictions for the agent to make. Such user teaching precisely addresses the limitations of the currently learned dialogue model, as these newly collected dialogue samples are driven by the agent's own policy. Specifically, in this study we let an expert user to correct the mistake made by the agent in tracking the user's goal at the end of each dialogue turn. This new batch of annotated dialogues are then added to the existing training corpus. We start the next round of supervised model training on this aggregated corpus to obtain an updated dialogue policy, and continue this dialogue imitation learning cycles.

    \begin{algorithm}[t]
    \caption{Dialogue Learning with Human Teaching and Feedback}
    \begin{algorithmic}[1]
    \State Train model end-to-end on dialogue samples $D$ with MLE and obtain policy $\pi _{\theta} (a|s)$ \Comment{eq 4}
    \For{learning iteration $k=1:K$}
        \State Run $\pi _{\theta} (a|s)$ with user to collect new \hspace*{5mm} dialogue samples $D _\pi$
        \State Ask user to correct the mistakes in the \hspace*{5mm} tracked user's goal for each dialogue turn \hspace*{5mm} in $D _\pi$
        \State Add the newly labeled dialogue samples \hspace*{5mm} to the existing corpora: $D \leftarrow D \cup D _\pi$
        \State Train model end-to-end on $D$ and obtain \hspace*{5mm} an updated policy $\pi _{\theta} (a|s)$ \Comment{eq 4}
    \EndFor
    \For{learning iteration $k=1:N$}
    	\State Run $\pi _{\theta} (a|s)$ with user for a new dialogue
        \State Collect user feedback as reward $r$
        \State Update model end-to-end and obtain an \hspace*{5mm} updated policy $\pi _{\theta} (a|s)$ \Comment{eq 5}
    \EndFor
    \end{algorithmic}
    \end{algorithm}

\subsection{Reinforcement Learning with Human Feedback}
    Learning from human teaching can be costly, as it requires expert users to provide corrections at each dialogue turn. We want to minimize the number of such imitation dialogue learning cycles and continue to improve the agent via a form of supervision signal that is easier to obtain. After the imitation learning stage, we further optimize the neural dialogue system with RL by letting the agent to interact with users and learn from user feedback. Different from the turn-level corrections in the imitation dialogue learning stage, the feedback is only collected at the end of a dialogue. A positive reward is collected for successful tasks, and a zero reward is collected for failed tasks. A step penalty is applied to each dialogue turn to encourage the agent to complete the task in fewer steps. In this work, we only use task-completion as the metric in designing the dialogue reward. One can extend it by introducing additional factors to the reward functions, such as naturalness of interactions or costs associated with KB queries. 
    
    To encourage the agent to explore the dialogue action space, we let the agent to follow a softmax policy during RL training by sampling system actions from the policy network outputs. We apply REINFORCE algorithm \cite{williams1992simple} in optimizing the network parameters. The objective function can be written as $J_k(\theta) = \mathbb E_{\theta}\left[ R_k  \right] = \mathbb E_{\theta}\left[ \sum_{t=0}^{K-k} \gamma ^{t}r_{k+t}  \right]$, with $\gamma \in [0,1)$ being the discount factor. With likelihood ratio gradient estimator, the gradient of the objective function can be derived as:
        \begin{equation}
            \begin{split}
            \nabla  _{\theta} J_k(\theta) &= \nabla _{\theta} \mathbb E_{\theta}\left[ R_k \right] \\
            &= \sum_{a_{k}}  \pi _{\theta}(a_{k} | s_{k}) \nabla _{\theta} \log \pi _{\theta}(a_{k} | s_{k}) R_{k} \\
            &= \mathbb E_{\theta}\left[ \nabla _{\theta} \log \pi _{\theta}(a_{k} | s_{k}) R_{k} \right]
            \end{split}
        \end{equation}
    This last expression above gives us an unbiased gradient estimator. 

\section{Experiments}
\subsection{Datasets}
    We evaluate the proposed method on DSTC2 \cite{henderson2014second} dataset in restaurant search domain and an internally collected dialogue corpus\footnote{The dataset can be accessed via \url{https://github.com/google-research-datasets/simulated-dialogue}.} in movie booking domain. The movie booking dialogue corpus has an average number of 8.4 turns per dialogue. Its training set has 100K dialogues, and the development set and test set each has 10K dialogues. 
    
    The movie booking dialogue corpus is generated~\cite{shah2018bootstrapping} using a finite state machine based dialogue agent and an agenda based user simulator~\cite{schatzmann2007agenda} with natural language utterances rewritten by real users. The user simulator can be configured with different personalities, showing various levels of randomness and cooperativeness. This user simulator is also used to interact with our end-to-end training agent during imitation and reinforcement learning stages. We randomly select a user profile when conducting each dialogue simulation. During model evaluation, we use an extended set of natural language surface forms over the ones used during training time to evaluate the generalization capability of the proposed end-to-end model in handling diverse natural language inputs.
    
\subsection{Training Settings}
    The size of the dialogue-level and utterance-level LSTM state is set as 200 and 150 respectively. Word embedding size is 300. Embedding size for system action and slot values is set as 32. Hidden layer size of the policy network is set as 100. We use Adam optimization method~\cite{kingma2014adam} with initial learning rate of 1e-3. Dropout rate of 0.5 is applied during supervised training to prevent the model from over-fitting. 
    
    In imitation learning, we perform mini-batch model update after collecting every 25 dialogues. System actions are sampled from the learned policy to encourage exploration. The system action is defined with the act and slot types from a dialogue act~\cite{henderson2013dialog}. For example, the dialogue act ``$confirm(date=monday)$'' is mapped to a system action ``$confirm\_date$'' and a candidate value ``$monday$'' for slot type ``$date$''. The slot types and values are from the dialogue state tracking output.
    
    In RL optimization, we update the model with every mini-batch of 25 samples. Dialogue is considered successful based on two conditions: (1) the goal slot values estimated from dialogue state tracking fully match to the user's true goal values, and (2) the system is able to confirm with the user the tracked goal values and offer an entity which is finally accepted by the user.  Maximum allowed number of dialogue turn is set as 15. A positive reward of +15.0 is given at the end of a successful dialogue, and a zero reward is given to a failed case. We apply a step penalty of -1.0 for each turn to encourage shorter dialogue for task completion. 

\subsection{Supervised Learning Results}
    Table \ref{tab:table_dstc2_sl} and Table \ref{tab:table_movie_corpus_sl} show the supervised learning model performance on DSTC2 and the movie booking corpus. Evaluation is made on DST accuracy. For the evaluation on DSTC2 corpus, we use the live ASR transcriptions as the user input utterances. Our proposed model achieves near state-of-the-art dialogue state tracking results on DSTC2 corpus, on both individual slot tracking and joint slot tracking, comparing to the recent published results using RNN~\cite{henderson2014robust} and neural belief tracker (NBT)~\cite{mrkvsic2016neural}. In the movie booking domain, our model also achieves promising performance on both individual slot tracking and joint slot tracking accuracy. Instead of using ASR hypothesis as model input as in DSTC2, here we use text based input which has much lower noise level in the evaluation of the movie booking tasks. This partially explains the higher DST accuracy in the movie booking domain comparing to DSTC2.

    \begin{table}[th]
      \caption{Dialogue state tracking results on DSTC2}
      \label{tab:table_dstc2_sl}
      \centering
      \begin{tabular}{l c c c c}
        \hline  
        \textbf{Model} & \textbf{Area}  & \textbf{Food}  & \textbf{Price}  & \textbf{Joint} \\
        \hline
        RNN & 92 & 86 & 86 & 69  \\
        RNN+sem. dict & 92 & 86 & 92 & 71  \\
        NBT & 90 & 84 & 94 & 72  \\
        Our SL model & 90 & 84 & 92 & 72  \\
        \hline
        \vspace*{-4ex}
      \end{tabular}     
    \end{table}

    \begin{table}[th]
      \caption{DST results on movie booking dataset}
      \label{tab:table_movie_corpus_sl}
      \centering
      \begin{tabular}{l c}
        \hline  
        \textbf{Goal slot} 		& \textbf{Accuracy} \\
        \hline  
        Num of Tickets		& 98.22 \\
        Movie 				& 91.86 \\
        Theater Name 		& 97.33 \\
        Date 				& 99.31 \\
        Time 				& 97.71 \\
        \hline
        Joint 				& 84.57 \\
        \hline
        \vspace*{-4ex}
      \end{tabular}     
    \end{table}
    
\subsection{Imitation and RL Results}
\label{sec:rl_results}
\begin{figure}[t]
          \centering
          \includegraphics[width=\linewidth]{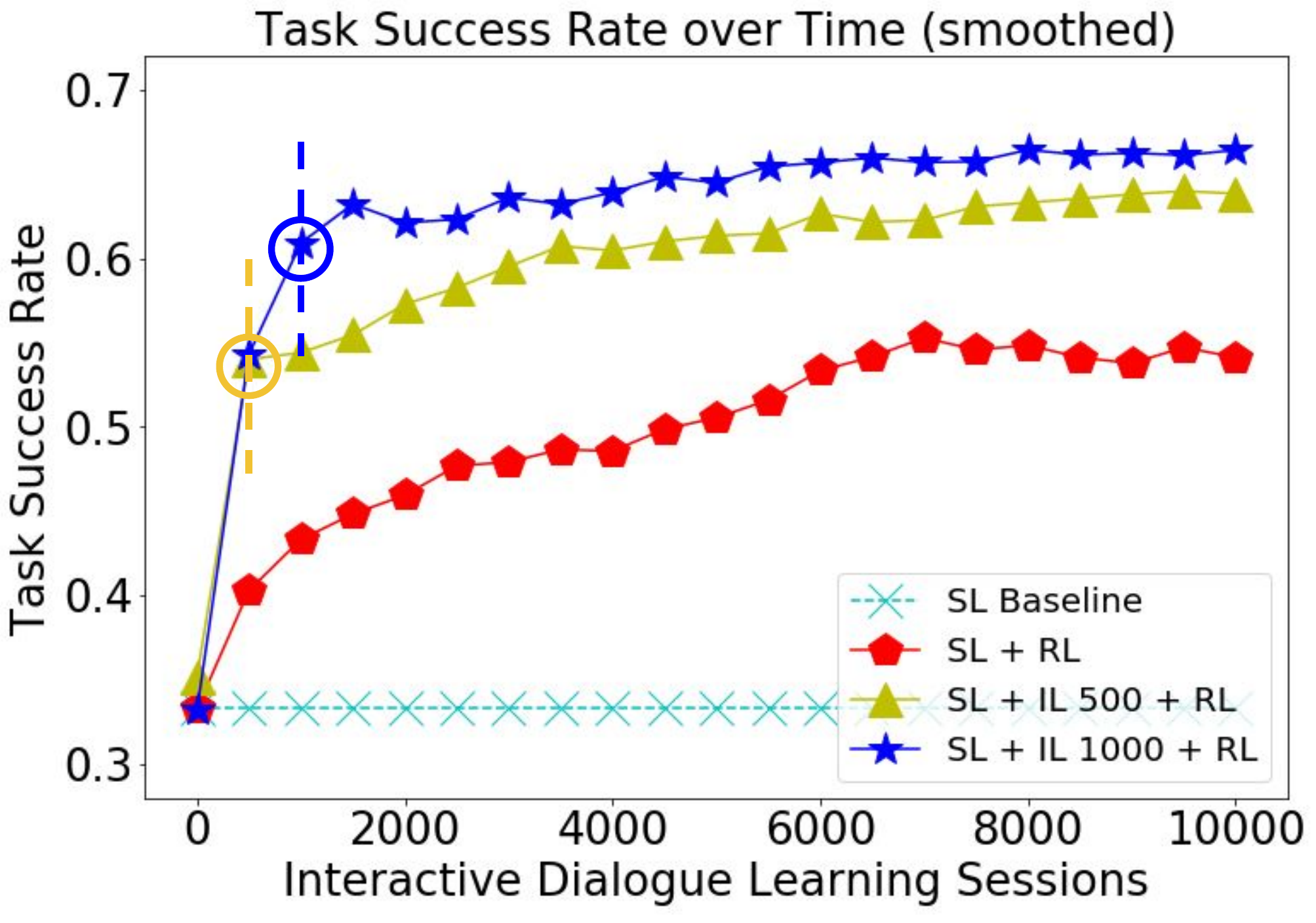}
          \vspace*{-4ex}
          \caption{Interactive learning curves on task success rate.}
          \label{fig:success_rate_curves}
        \end{figure}

\begin{figure}[t]
          \centering
          \includegraphics[width=\linewidth]{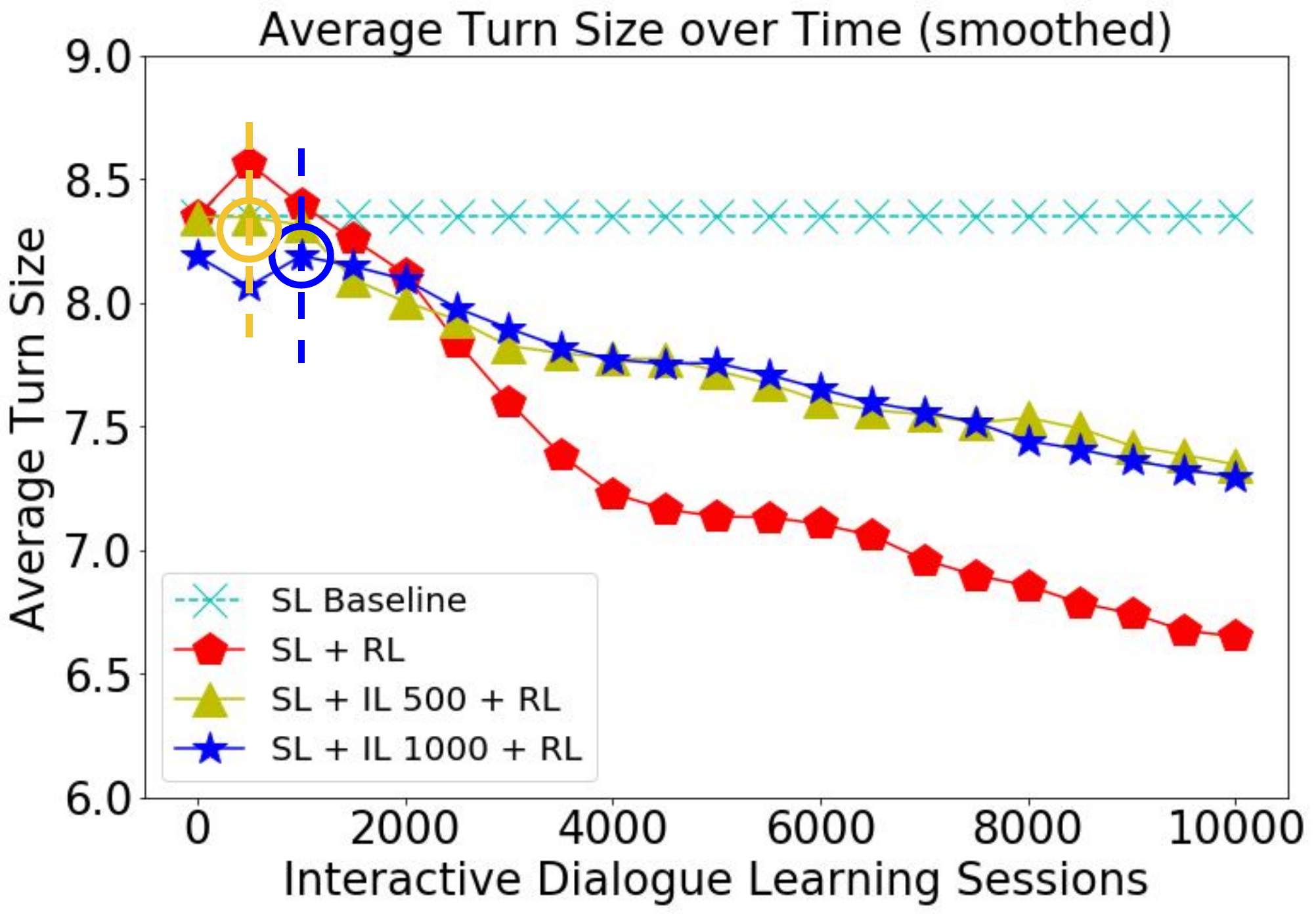}
          \vspace*{-4ex}
          \caption{Interactive learning curves on average dialogue turn size.}
          \label{fig:turn_size_curves}
        \end{figure}

\begin{figure}[t]
          \centering
          \includegraphics[width=\linewidth]{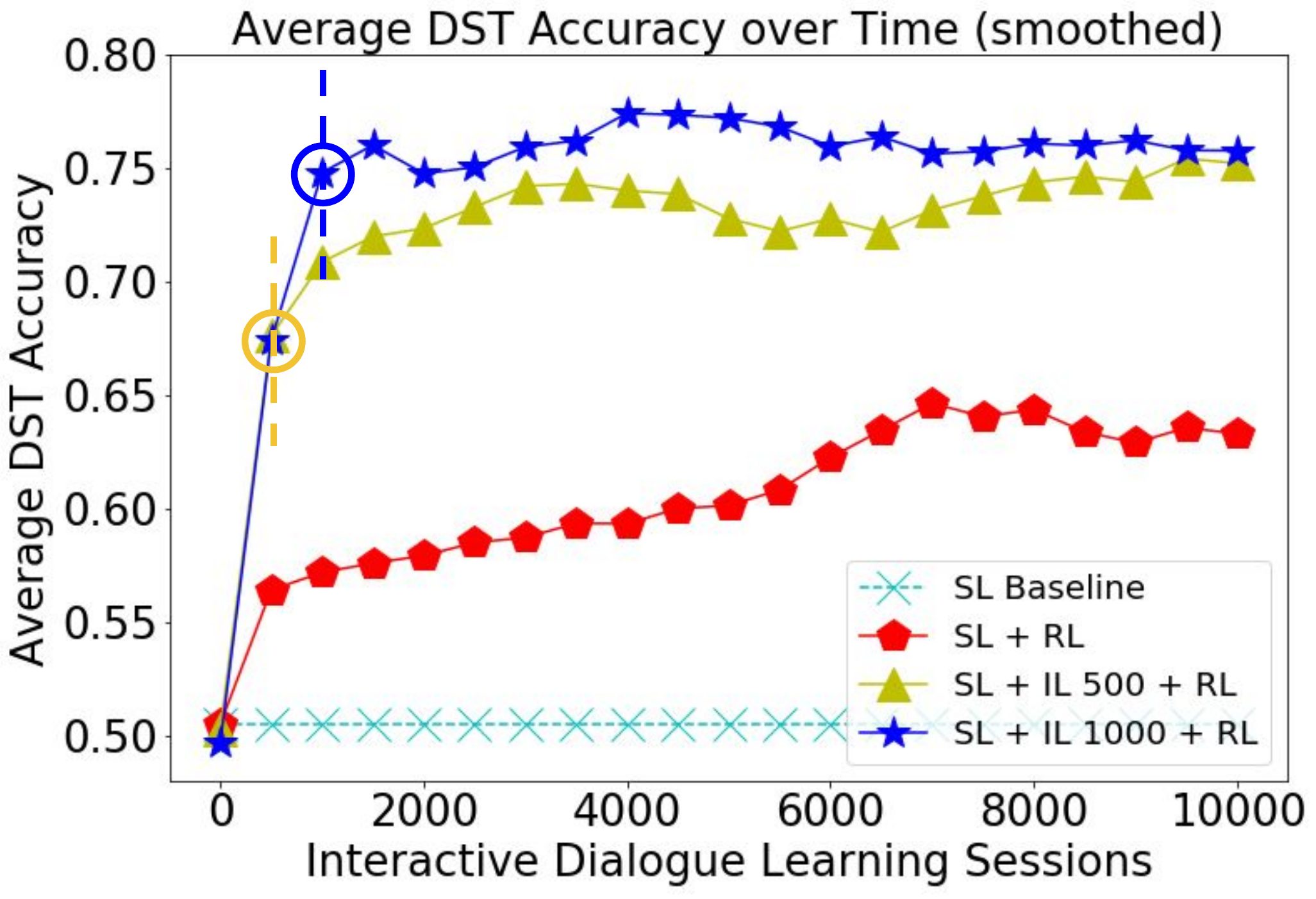}
          \vspace*{-4ex}
          \caption{Interactive learning curves on dialogue state tracking accuracy.}
          \label{fig:dst_curves}
        \end{figure}

    Evaluations of interactive learning with imitation and reinforcement learning are made on metrics of (1) task success rate, (2) dialogue turn size, and (3) DST accuracy. Figures \ref{fig:success_rate_curves}, \ref{fig:turn_size_curves}, and \ref{fig:dst_curves} show the learning curves for the three evaluation metrics. In addition, we compare model performance on task success rate using two different RL training settings, the end-to-end training and the policy-only training, to show the advantages of performing end-to-end system optimization with RL.

    \textbf{Task Success Rate} \hspace{3mm} As shown in the learning curves in Figure \ref{fig:success_rate_curves}, the SL model performs poorly. This might largely due to the compounding errors caused by the mismatch of dialogue state distribution between offline training and interactive learning. We use an extended set of user NLG templates during interactive evaluation. Many of the test NLG templates are not seen by the supervised training agent. Any mistake made by the agent in understanding the user's request may lead to compounding errors in the following dialogue turns, which cause final task failure. The red curve ({\tt SL + RL}) shows the performance of the model that has RL applied on the supervised pre-training model. We can see that interactive learning with RL using a weak form of supervision from user feedback continuously improves the task success rate with the growing number of user interactions. We further conduct experiments in learning dialogue model from scratch using only RL (i.e. without supervised pre-training), and the task success rate remains at a very low level after 10K dialogue simulations. We believe that it is because the dialogue state space is too complex for the agent to learn from scratch, as it has to learn a good NLU model in combination with a good policy to complete the task. The yellow curve ({\tt SL + IL 500 + RL}) shows the performance of the model that has 500 episodes of imitation learning over the SL model and continues with RL optimization. It is clear from the results that applying imitation learning on supervised training model efficiently improves task success rate. RL optimization after imitation learning increases the task success rate further. The blue curve ({\tt SL + IL 1000 + RL}) shows the performance of the model that has 1000 episodes of imitation learning over the SL model and continues with RL. Similarly, it shows hints that imitation learning may effectively adapt the supervised training model to the dialogue state distribution during user interactions. 
    
    \textbf{Average Dialogue Turn Size} \hspace{3mm} Figure \ref{fig:turn_size_curves} shows the curves for the average turn size of successful dialogues. We observe decreasing number of dialogue turns in completing a task along the growing number of interactive learning sessions. This shows that the dialogue agent learns better strategies in successfully completing the task with fewer number of dialogue turns. The red curve with RL applied directly after supervised pre-training model gives the lowest average number of turns at the end of the interactive learning cycles, comparing to models with imitation dialogue learning. This seems to be contrary to our observation in Figure \ref{fig:success_rate_curves} that imitation learning with human teaching helps in achieving higher task success rate. By looking into the generated dialogues, we find that the {\tt SL + RL} model can handle easy tasks well but fails to complete more challenging tasks. Such easy tasks typically can be handled with fewer number of turns, which result in the low average turn size for the {\tt SL + RL} model. On the other hand, the imitation plus RL models attempt to learn better strategies to handle those more challenging tasks, resulting in higher task success rates and also slightly increased dialogue length comparing to {\tt SL + RL} model.

    \textbf{Dialogue State Tracking Accuracy} \hspace{3mm} Similar to the results on task success rate, we see that imitation learning with human teaching quickly improves dialogue state tracking accuracy in just a few hundred interactive learning sessions. The joint slots tracking accuracy in the evaluation of SL model using fixed corpus is 84.57\% as in Table \ref{tab:table_movie_corpus_sl}. The accuracy drops to 50.51\% in the interactive evaluation with the introduction of new NLG templates. Imitation learning with human teaching effectively adapts the neural dialogue model to the new user input and dialogue state distributions, improving the DST accuracy to 67.47\% after only 500 imitation dialogue learning sessions. Another encouraging observation is that RL on top of SL model and IL model not only improves task success rate by optimizing dialogue policy, but also further improves dialogue state tracking performance. This shows the benefits of performing end-to-end optimization of the neural dialogue model with RL during interactive learning. 

\textbf{End-to-End RL Optimization} 
        \begin{figure}[t]
          \centering
          \includegraphics[width=\linewidth]{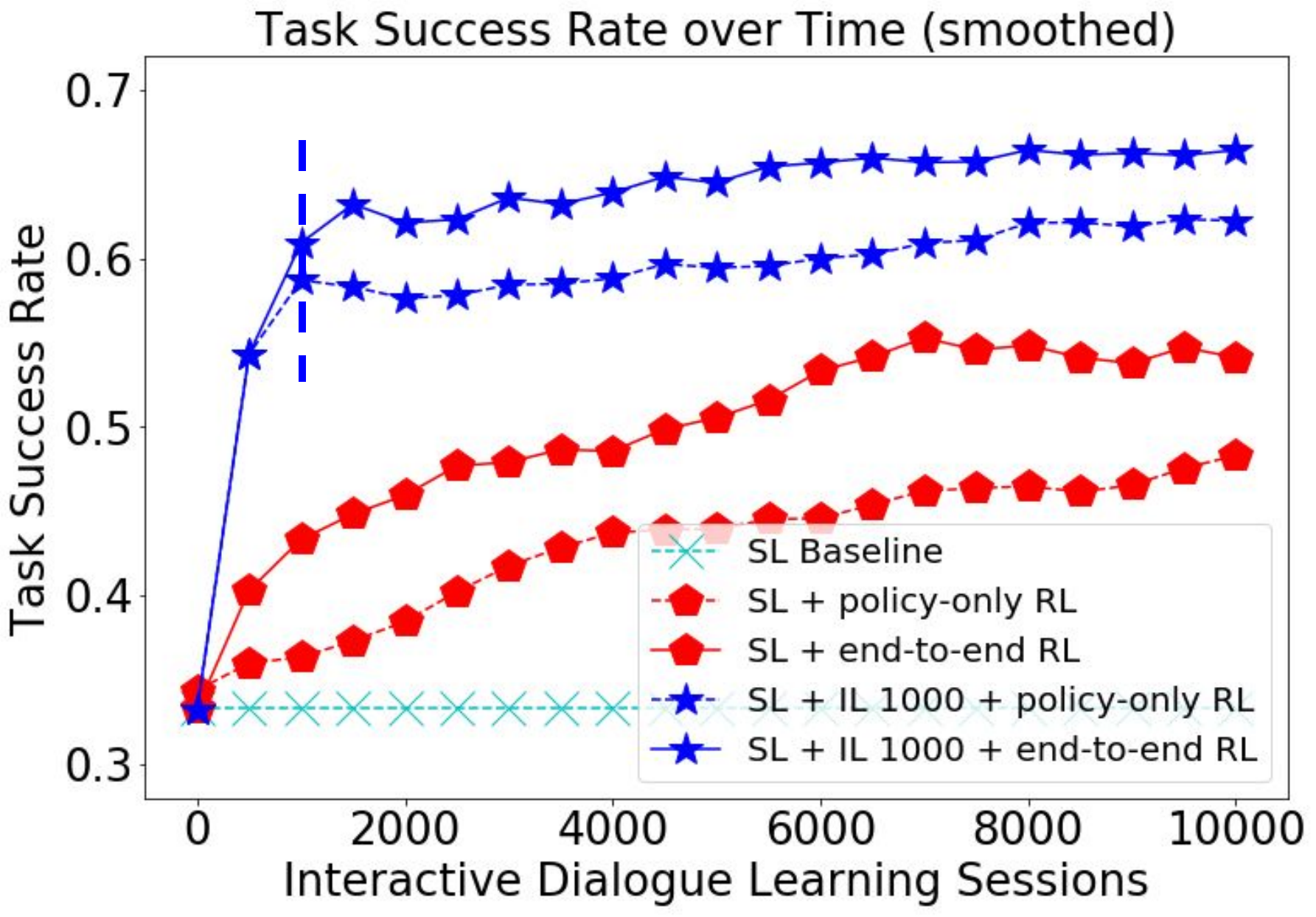}
          \vspace*{-4ex}
          \caption{Interactive learning curves on task success rate with different RL training settings.}
          \label{fig:success_rate_e2e_update}
        \end{figure}
    To further show the benefit of performing end-to-end optimization of dialogue agent, we compare models with two different RL training settings, the end-to-end training and the policy-only training. End-to-end RL training is what we applied in previous evaluation sections, in which the gradient propagates from system action output layer all the way back to the natural language user input layer. Policy-only training refers to only updating the policy network parameters during interactive learning with RL, with all the other underlying system parameters fixed. The evaluation results are shown in Figure \ref{fig:success_rate_e2e_update}. From these learning curves, we see clear advantage of performing end-to-end model update in achieving higher dialogue task success rate during interactive learning comparing to only updating the policy network. 

\subsection{Human User Evaluations}
\label{sec:human_eval_results}
    We further evaluate the proposed method with human judges recruited via Amazon Mechanical Turk. Each judge is asked to read a dialogue between our model and user simulator and rate each system turn on a scale of 1 (frustrating) to 5 (optimal way to help the user). Each turn is rated by 3 different judges. We collect and rate 100 dialogues for each of the three models: (i) SL model, (ii) SL model followed by 1000 episodes of IL, (iii) SL and IL followed by RL. Table \ref{tab:eval_result_human} lists the mean and standard deviation of human scores overall system turns. Performing interactive learning with imitation and reinforcement learning clearly improves the quality of the model according to human judges.
    
    \begin{table}[th]
    \centering
    \begin{tabular}{l|c}
    \hline
    \textbf{Model} & \textbf{Score}      \\ \hline
    SL & 3.987 $\pm$ 0.086  \\
    SL + IL 1000 & 4.378 $\pm$ 0.082 \\
    SL + IL 1000 + RL & 4.603 $\pm$ 0.067 \\ \hline
    \end{tabular}
    \caption{Human evaluation results. Mean and standard deviation of crowd worker scores (between 1 to 5).}
    \label{tab:eval_result_human}
    \vspace*{-1ex}
    \end{table}

\section{Conclusions}
    In this work, we focus on training task-oriented dialogue systems through user interactions, where the agent improves through communicating with users and learning from the mistake it makes. We propose a hybrid learning approach for such systems using end-to-end trainable neural network model. We present a hybrid imitation and reinforcement learning method, where we firstly train a dialogue agent in a supervised manner by learning from dialogue corpora, and continuously to improve it by learning from user teaching and feedback with imitation and reinforcement learning. We evaluate the proposed learning method with both offline evaluation on fixed dialogue corpora and interactive evaluation with users. Experimental results show that the proposed neural dialogue agent can effectively learn from user teaching and improve task success rate with imitation learning. Applying reinforcement learning with user feedback after imitation learning with user teaching improves the model performance further, not only on the dialogue policy but also on the dialogue state tracking in the end-to-end training framework.

\bibliography{naaclhlt2018}
\bibliographystyle{acl_natbib}

\appendix

\end{document}